\documentclass{article}

\usepackage[english]{babel}

\usepackage[letterpaper,top=2cm,bottom=2cm,left=3cm,right=3cm,marginparwidth=1.75cm]{geometry}

\usepackage{amsmath}
\usepackage{graphicx}
\usepackage{float}
\usepackage[colorlinks=true, allcolors=blue]{hyperref}
\usepackage{authblk}  
\usepackage{booktabs}
\usepackage{amssymb,amsfonts}
\usepackage{todonotes}
\usepackage{comment}
\usepackage{adjustbox} 
\usepackage[
    capitalise,
    noabbrev,
  ]{cleveref}
\usepackage{bm}
\usepackage[labelfont=bf]{caption}

\newcommand{\bnabla}{\bm{\nabla}} 
\newcommand{\TV}{\mathrm{TV}}

\title{FractalPINN-Flow: A Fractal-Inspired Network for Unsupervised Optical Flow Estimation with Total Variation Regularization}

\author[1]{Sara Behnamian}
\author[2]{Rasoul Khaksarinezhad}
\author[3]{Andreas Langer}

\affil[1]{Globe Institute, University of Copenhagen, Øster Voldgade 5-7, 1350 Copenhagen K, Denmark\\ \texttt{sara.behnamian@sund.ku.dk}}
\affil[2]{\texttt{rasoul.khaksari@gmail.com}}
\affil[3]{Centre for Mathematical Sciences, Lund University, Box 118, 221 00 Lund, Sweden\\ \texttt{andreas.langer@math.lth.se}}

\date{}

\begin{document}

\maketitle

\begin{abstract}
We present FractalPINN-Flow, an unsupervised deep learning framework for dense optical flow estimation that learns directly from consecutive grayscale frames without requiring ground truth. The architecture centers on the Fractal Deformation Network (FDN)—a recursive encoder-decoder inspired by fractal geometry and self-similarity. Unlike traditional CNNs with sequential downsampling, FDN uses repeated encoder-decoder nesting with skip connections to capture both fine-grained details and long-range motion patterns.

The training objective is based on a classical variational formulation using total variation (TV) regularization. Specifically, we minimize an energy functional that combines $L^1$ and $L^2$ data fidelity terms to enforce brightness constancy, along with a TV term that promotes spatial smoothness and coherent flow fields.

Experiments on synthetic and benchmark datasets show that FractalPINN-Flow produces accurate, smooth, and edge-preserving optical flow fields. The model is especially effective for high-resolution data and scenarios with limited annotations.
\end{abstract}


\section{Introduction}

Optical flow estimation seeks to recover the apparent motion field between two consecutive grayscale images. Let \( I_1, I_2 : \Omega \to [0,1] \) denote two discrete images defined on a spatial domain \( \Omega \subset \mathbb{Z}^2 \), and let \( w : \Omega \to \mathbb{R}^2 \), \( w(x) = (u(x), v(x)) \), be the displacement field to be estimated. The fundamental assumption underlying most optical flow methods is brightness constancy, which posits that the intensity of each point remains constant as it moves, i.e.,
\begin{equation}\label{Eq:ConstantBrightness}
    I_1(x) = I_2(x + w(x)) \quad \text{for all } x \in \Omega.
\end{equation}
This results in a nonlinear constraint on \( w \). Since the problem is underdetermined (one equation, two unknowns), additional regularization is required.

To simplify the nonlinear data term, many methods approximate the brightness constancy relation \eqref{Eq:ConstantBrightness} by a linear model:
\[
I_2(x + w(x)) \approx I_2(x) + \nabla I_2(x) \cdot w(x),
\]
where \( \nabla I_2 \) denotes the image gradient. This yields the approximate linear constraint
\[
\nabla I_2(x) \cdot w(x) + I_2(x) - I_1(x) = 0,
\]
as used in the classical Horn--Schunck formulation~\cite{horn1981determining}. There the flow field  is recovered by minimizing 
\[
\|\nabla I_2 \cdot w + I_2 - I_1\|_2^2 + \lambda \|\bnabla w\|_2^2, 
\]
where \( \lambda > 0 \) is a regularization parameter, \( \| \bnabla w \|_2^2 = \| \nabla u \|_2^2 + \| \nabla v \|_2^2 \) encourages global smoothness of the displacement field and $\|\cdot\|_2$ denotes the standard Euclidean norm. This quadratic regularization penalizes rapid variations in the flow but tends to oversmooth motion boundaries.

A more robust alternative replaces the $L^2$-norm with the $L^1$-norm in both the data fidelity and smoothness terms, yielding a total variation (TV) regularization model, allowing the flow field to exhibit discontinuities. This leads to the energy
\begin{equation*}
\|\nabla I_2 \cdot w + I_2 - I_1\|_1 + \lambda \|\bnabla w\|_1, 
\end{equation*}
where the last term represents the total variation of the flow field $w$. 

In practice, this formulation encourages piecewise-smooth flow fields while preserving motion discontinuities~\cite{zach2007duality}.
To efficiently solve TV-regularized optical flow problems, in~\cite{Hilb2023} a primal-dual finite element method was introduced and combined with an iterative warping algorithm to handle large displacements. Building on this in \cite{Alkaemper2025,Jacumin2025} adaptive discretizations schemes were proposed, which can be interpreted as an adaptive multigrid method, further accelerating convergence and sometimes even improving the quality of the estimated flow field.

With the advent of deep learning, optical flow estimation has seen dramatic advances. FlowNet~\cite{dosovitskiy2015flownet} introduced a convolutional neural network (CNN) architecture capable of predicting flow directly from image pairs, opening the door to data-driven solutions, see also~\cite{shi2023}. However, supervised methods require large-scale annotated datasets, which are difficult to obtain for many real-world domains.
To address this challenge, unsupervised approaches have gained momentum by optimizing photometric consistency and smoothness priors without ground-truth supervision. These methods can be further classified according to the number of input frames used during training, including multiframe models~\cite{stone2021}, which leverage temporal consistency across sequences, and two-frame models~\cite{jonschkowski2020,meister2018}, which are based solely on image pairs. Although most of such models incorporate smoothness losses, they typically do not employ TV regularization explicitly.

In this paper, we propose \textit{FractalPINN-Flow}, a novel unsupervised framework that integrates an encoding-decoding strategy with total variation regularization to estimate dense optical flow from grayscale image sequences. Our use of TV minimization in a neural setting is motivated by recent work such as DeepTV~\cite{behnamian2024deeptv}, which demonstrated how classical regularization techniques can be effectively combined with deep architectures to produce sharp, structurally coherent outputs. At the core of our model is a \textit{Fractal Deformation Network (FDN)}, a recursive encoder-decoder architecture inspired by the self-similar principles of FractalNet~\cite{larsson2016fractalnet}. Unlike conventional hierarchical CNNs, which apply sequential downsampling, the FDN recursively nests encoder-decoder modules at multiple scales, each equipped with skip connections. This design builds a deep multiscale representation that maintains local texture while capturing long-range deformation structures, properties that are especially important in scenes with fine motion patterns or limited training data.
The FDN output is passed to a lightweight CNN that predicts the optical flow field. Training is fully unsupervised, uses only two frames and is guided by a composite loss function that consists of a combined $L^1$/$L^2$ data term and TV regularization.

\section{FractalPINN-Flow: Architecture and Implementation}

We consider the problem of learning the optical flow from two input images $I_1$, $I_2$ by minimizing 
\begin{equation}\label{Eq:L1L2TV}
E_{\TV}(w) := \lambda_1 \|\nabla I_2 \cdot w + I_2 - I_1\|_1+\lambda_2 \|\nabla I_2 \cdot w + I_2 - I_1\|_2^2 + \lambda_\TV \|\bnabla w\|_1,
\end{equation}
where $\lambda_1,\lambda_2,\lambda_\TV \ge 0$, $w$ is represented by a neural network, see \cref{Sec:NN} for a detailed description of the network structure, and 
\begin{equation}\label{Eq:aTV}
    \|\bnabla w\|_1 := \| \nabla_x u \|_1 +  \| \nabla_y u \|_1 +\| \nabla_x v \|_1 +  \| \nabla_y v \|_1
\end{equation} 
is the anisotropic TV. Here, $\nabla_x$ and $\nabla_y$ denote the finite differences along the horizontal and vertical axes, respectively. This regularization penalizes abrupt discontinuities while preserving meaningful motion boundaries, thereby promoting spatially coherent and piecewise-smooth flow estimates. We note that while the gradients in \eqref{Eq:aTV} could also be implemented in a pointwise manner, as is common in physics-informed neural networks \cite{RaPeKa:19}, we refrain from doing so following the rationale in~\cite{behnamian2024deeptv}. In particular, pointwise evaluation may completely miss jump discontinuities, especially in piecewise constant outputs, by sampling points where the gradient happens to vanish. In contrast, a finite difference discretization reliably captures such variations and provides a more accurate measure of total variation.

We note that a combined \( L^1 \)/\( L^2 \) data fidelity term together with total variation regularization, as used in~\eqref{Eq:L1L2TV}, was first introduced in~\cite{Hintermuller2013} and analyzed in~\cite{Langer:17a} in the context of variational image restoration.  We consider this formulation here for optical flow estimation.

\subsection{Neural Network}\label{Sec:NN}

We propose a deep learning architecture for dense optical flow estimation that integrates our novel FDN with a CNN-based optical flow prediction head. The method estimates pixel-wise motion between two consecutive frames by leveraging multi-scale fractal features.

\subsubsection{Network Architecture}
\paragraph{High-Level Structure.} The architecture of the neural networks we use follows a sequential flow of processing stages, outlined as follows:

\begin{center}
\begin{tabular}{c}
\textbf{Input:} Two consecutive grayscale frames $(I_1, I_2)$ \\
$\downarrow$ \\
\textbf{FDN:} Encoder-decoder network \\
$\downarrow$ \\
\textbf{Projection Layer:} $32 \rightarrow 64$ channels \\
$\downarrow$ \\
\textbf{Optical Flow Network:} CNN with 5 layers \\
$\downarrow$ \\
\textbf{Output:} Dense optical flow field $(u, v)$
\end{tabular}
\end{center}

The FDN encodes multi-scale motion features, which are then mapped to dense flow fields by a convolutional regression head. This modular structure mirrors successful encoder-decoder designs widely used in optical flow~\cite{sun2018pwc, teed2020raft} and image registration~\cite{balakrishnan2019voxelmorph}.

\paragraph{Fractal Deformation Network (FDN).}
The FDN is based on a symmetric U-Net-style encoder–decoder architecture with configurable depth $d$. In this work, we fix the depth to $d = 4$ without hyperparameter tuning. Each downsampling (encoder) block consists of two $3 \times 3$ convolutional layers, each followed by batch normalization and ReLU activation, and concludes with a $2 \times 2$ max pooling operation. For $d = 4$, the channel configuration expands as  
\begin{center}
\texttt{2 $\rightarrow$ 32 $\rightarrow$ 64 $\rightarrow$ 128 $\rightarrow$ 256}
\end{center}
enabling progressive abstraction of features at increasingly coarser spatial resolutions.

The overall architecture is designed to promote multi-scale feature extraction through this hierarchical structure. Although we refer to it as a Fractal Deformation Network for consistency with the naming of our framework, the term "fractal" here is used loosely to suggest repeated block-level processing across scales, rather than strict self-similarity or recursion. This design allows the network to capture both fine-scale deformations and large displacements efficiently.

The decoder mirrors the encoder and progressively increases spatial resolution using $2 \times 2$ transposed convolutions. These operations serve as learned upsampling layers: they first insert zeros between pixels (to increase spatial resolution) and then apply a learnable $2\times 2$ kernel to interpolate meaningful values, while also reducing the number of channels by half (e.g., $256 \to 128$). The overall flow of channel dimensions of the decoder is given by: 
\begin{center}
    \texttt{256 \(\rightarrow\) 128 \(\rightarrow\) 64 \(\rightarrow\) 32 \(\rightarrow\) 32.}
\end{center} 
To preserve high-resolution details, skip connections are added from the encoder to decoder at each intermediate resolution level (excluding the input and final output layers, as their channel dimensions differ and cannot be connected directly). These connections use bilinear interpolation of the encoder feature maps to match the size of the decoder features, followed by element-wise addition. After this fusion, each decoder block includes two $3 \times 3$ convolutions, each followed by batch normalization and ReLU activation, to refine the combined features.

To ensure compatibility with the downsampling operations, the input images are zero-padded, maintaining consistent spatial dimensions throughout the network.
Unlike classical U-Net models that concatenate features, our architecture uses element-wise addition instead of concatenation, preserving multiscale information while reducing the number of trainable parameters. The final output of the decoder and thus the FDN is a 32-channel feature map matching the input spatial size, which is passed to the subsequent module.  This higher dimension (32-channel) representation
allows the network to encode more complex information about the image differences and potential motion cues than a lower-dimensional (e.g., 2-channel) feature representation could encode.

\paragraph{Projection Layer}
To adapt the FDN output to the optical flow predictor, a $1\times1$ convolution projects 32 channels to 64. This shallow transformation learns to emphasize motion-relevant features while maintaining full spatial resolution. It ensures architectural compatibility without adding significant computational burden.

\paragraph{Optical Flow Prediction Network}
This subnetwork is a compact CNN that predicts dense flow fields. It consists of five $3\times3$ convolutional layers with ReLU activations and no pooling. The channel pattern follows:
\begin{center}
\texttt{64 $\rightarrow$ 128 $\rightarrow$ 256 $\rightarrow$ 128 $\rightarrow$ 64 $\rightarrow$ 2}
\end{center}
No downsampling is performed, preserving exact spatial correspondence between features and motion vectors. The final layer has two output channels without ReLU activations, representing horizontal and vertical displacement per pixel. The use of a bottleneck architecture enables expressive mapping while retaining efficiency.

\paragraph{End-to-End Information Flow}
The full architecture forms a feedforward pipeline:
\begin{align*}
    \text{Input Pair: } & I = [I_1, I_2] \in \mathbb{R}^{B \times 2 \times H \times W} \\
    \text{FDN Output: } & F \in \mathbb{R}^{B \times 32 \times H \times W} \\
    \text{Projected Features: } & P \in \mathbb{R}^{B \times 64 \times H \times W} \\
    \text{Predicted Flow: } & w \in \mathbb{R}^{B \times 2 \times H \times W}
\end{align*}
where $B$ denotes the batch size and $H\times W$ the number of pixels in the images $I_1$ and $I_2$.
This separation of concerns ensures that multiscale feature learning and dense motion regression are individually optimized yet trained jointly.

\section{Numerical Evaluation}

\subsection{Evaluation Metrics}

\paragraph{Average Endpoint Error (AEE)}
The AEE metric quantifies the Euclidean distance between the predicted and ground truth flow vectors across all pixels:

\begin{equation*}
\text{AEE} = \frac{1}{N} \sum_{i=1}^{N} \left\| w_{\text{pred}}^{(i)} - w_{\text{gt}}^{(i)} \right\|_2,
\end{equation*}

\noindent where $w_{\text{pred}}^{(i)}$ and $w_{\text{gt}}^{(i)}$ denote the $i$-th pixel of the predicted and ground truth flow vectors $w_{\text{pred}}$ and $w_{\text{gt}}$, respectively, and $N$ is the number of pixels. 

\paragraph{Average Angular Error (AAE)}
To assess directional consistency, the AAE metric measures angular discrepancies between predicted and ground truth flow vectors:

\begin{align*}
\text{AAE} = \frac{1}{N} \sum_{i=1}^{N} \arccos \left( \cos(\theta^{(i)}) \right), \quad \cos(\theta) = \frac{u_{\text{pred}} u_{\text{gt}} + v_{\text{pred}} v_{\text{gt}} + 1}{\sqrt{(u_{\text{pred}}^2 + v_{\text{pred}}^2 + 1)(u_{\text{gt}}^2 + v_{\text{gt}}^2 + 1)}}.
\end{align*}

\noindent The additive constant term in both the numerator and denominator ensures numerical stability, particularly in zero-flow regions, and prevents division by zero.
\subsection{Training Configuration}

To evaluate the impact of regularization techniques on optical flow estimation, we optimize our functional~\eqref{Eq:L1L2TV} with $\lambda_1 = 0.2$, $\lambda_2 = 0.8$, and varying TV weights $\lambda_{\mathrm{TV}}$. All experiments are conducted on a synthetic image and the Middlebury dataset using PyTorch~\cite{paszke2019pytorch}, with deterministic settings to ensure reproducibility. Logging and visualization are automated for all runs.

In all configurations, we use the Adam optimizer~\cite{KingmaBa:15} with a learning rate of $10^{-4}$. Training is conducted for a fixed number of epochs, which serves as the stopping criterion. Each input consists of a pair of normalized grayscale frames, concatenated along the channel dimension. Since we use only one image pair—two frames $I_1$ and $I_2$—for training, the batch size is set to 1.

We employ the FDN with a fixed depth of $d = 4$. Although the architecture supports configurable depth, we do not perform hyperparameter tuning on this parameter; all models use the same setting to ensure comparability. This depth balances model capacity and computational efficiency.

The neural network selection is based on the training loss, and the best model checkpoint (i.e., with the lowest loss) is saved for each configuration. At each epoch, we compute the data and regularization terms, along with endpoint and angular error metrics: average endpoint error (AEE), standard deviation of endpoint error (SDEE), average angular error (AAE), and standard deviation of angular error (SDAE). These metrics are used to evaluate convergence behavior and to guide the final result visualization and comparison.

\paragraph{Infrastructure and Logging}

All experiments have been executed on NVIDIA GPUs using PyTorch with CUDA acceleration. The training framework supports comprehensive experiment tracking through structured logging. For each configuration, the system creates a dedicated directory containing JSON-formatted configuration files, full training logs, and performance summaries.

Loss curves and evaluation metrics (AEE, AAE, and their standard deviations) are recorded at every epoch.  To maintain memory stability during long training runs, the system applies aggressive memory management strategies. After each epoch, the CUDA cache is explicitly cleared, and Python's garbage collector is invoked to prevent memory accumulation. This ensures robustness when running multiple configurations sequentially on large datasets.

\subsection{Shepp-Logan Phantom with Synthetic Motion}

To validate our method under idealized and interpretable conditions, we construct a controlled synthetic experiment based on the Shepp-Logan phantom~\cite{shepp1974fourier}—a canonical analytic image commonly used in medical imaging and tomographic reconstruction due to its smooth grayscale transitions and well-defined elliptical structures. The phantom is set to $256 \times 256$ pixels and augmented with two distinct circular regions 
to simulate localized anatomical structures. The first circle, with moderate intensity (0.5), is positioned in the upper quadrant, while the second circle, with higher intensity (0.75), is placed in the lower quadrant. These intensity values create sufficient contrast against the phantom background while maintaining realistic tissue-like appearance.

\begin{figure}[h]
    \centering
    \includegraphics[width=0.23\textwidth]{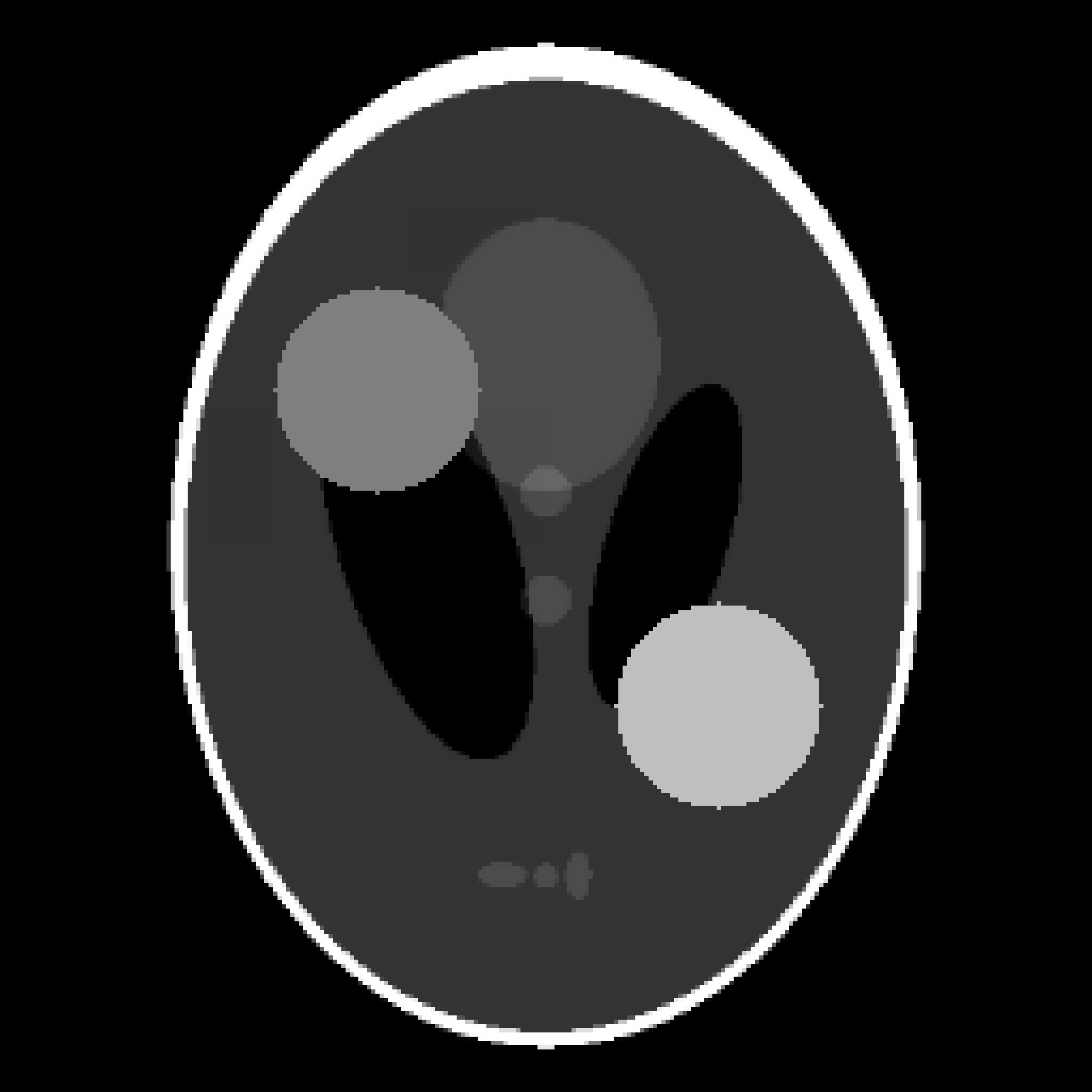}
    \includegraphics[width=0.23\textwidth]{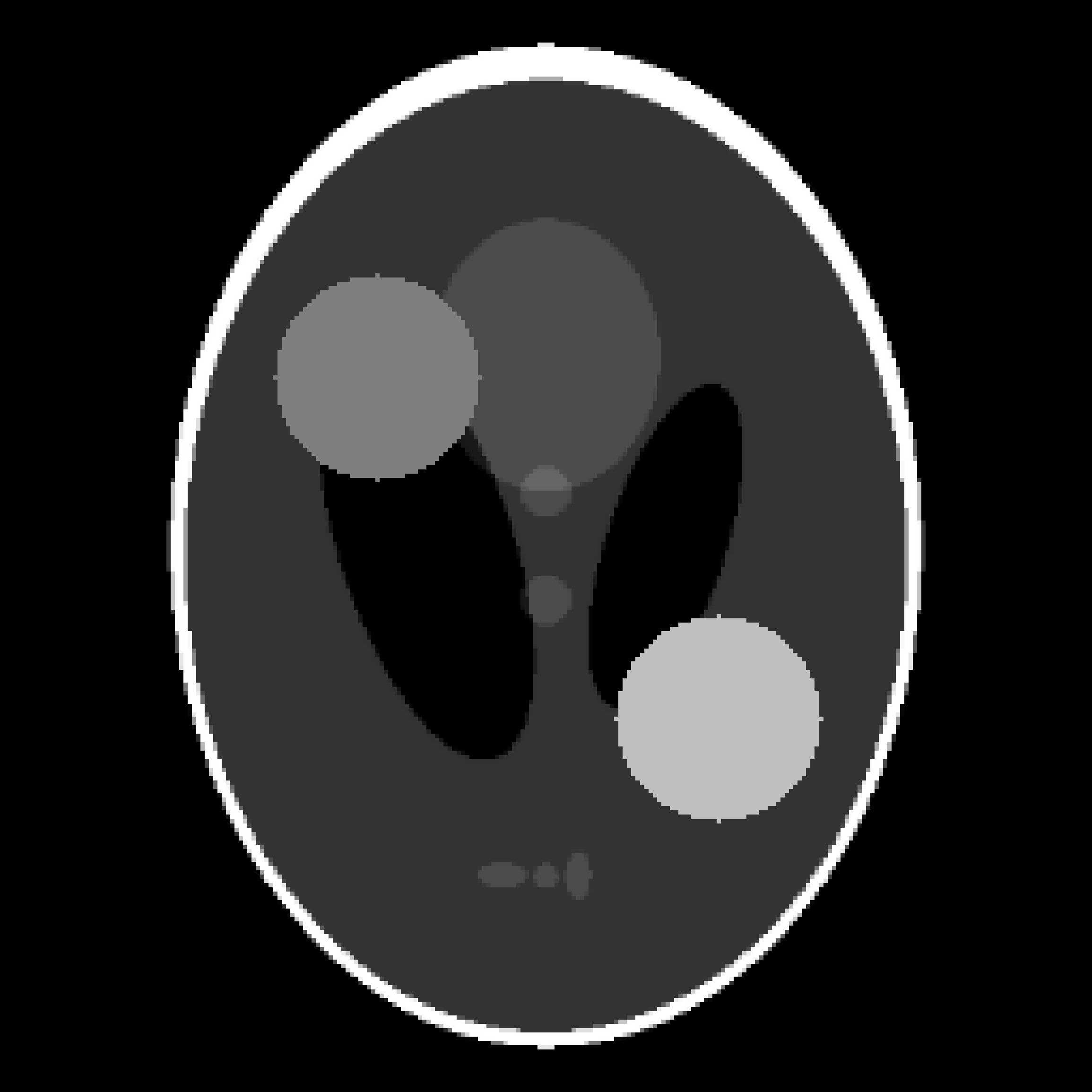}
    \includegraphics[width=0.23\textwidth]{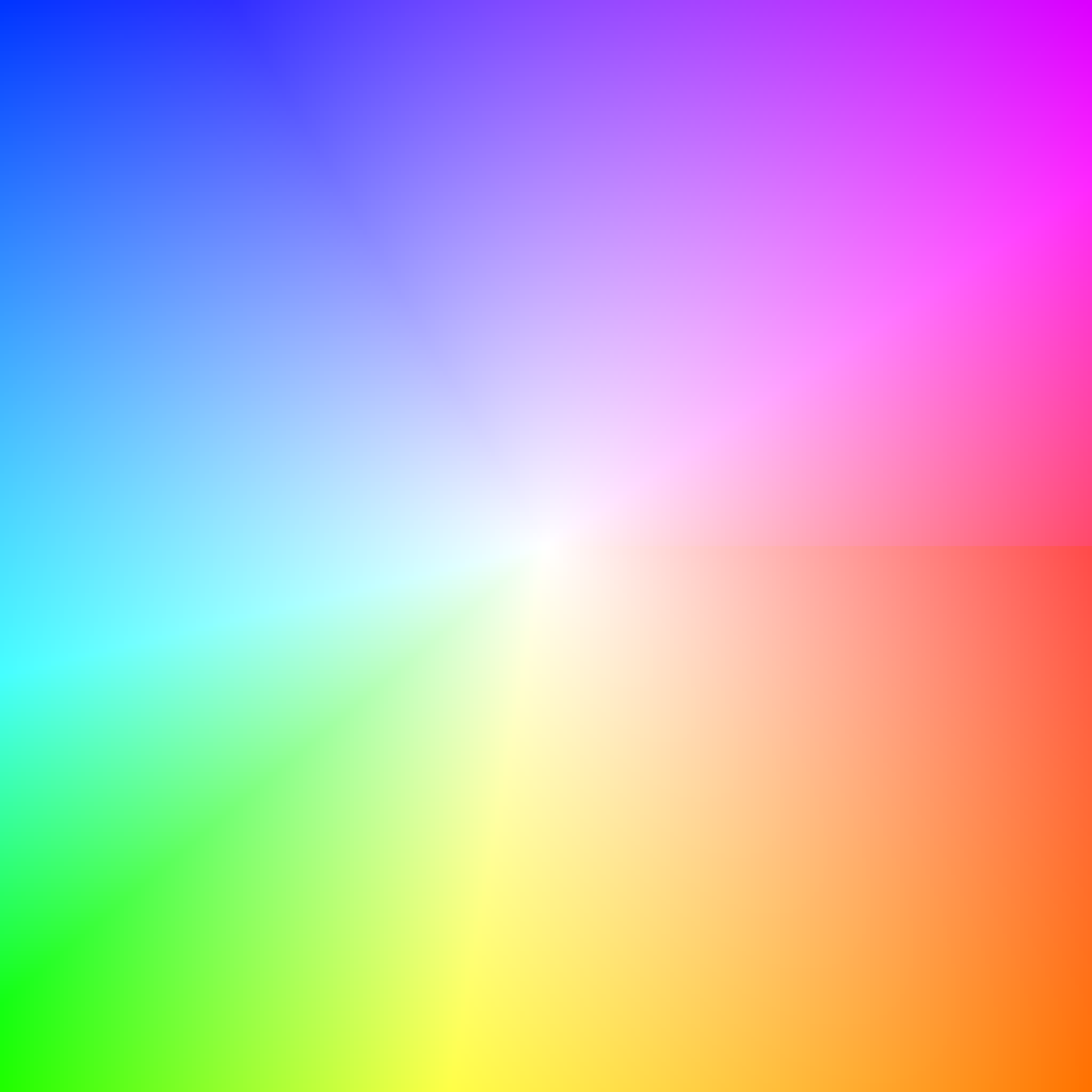}\\
    \includegraphics[width=0.23\textwidth]{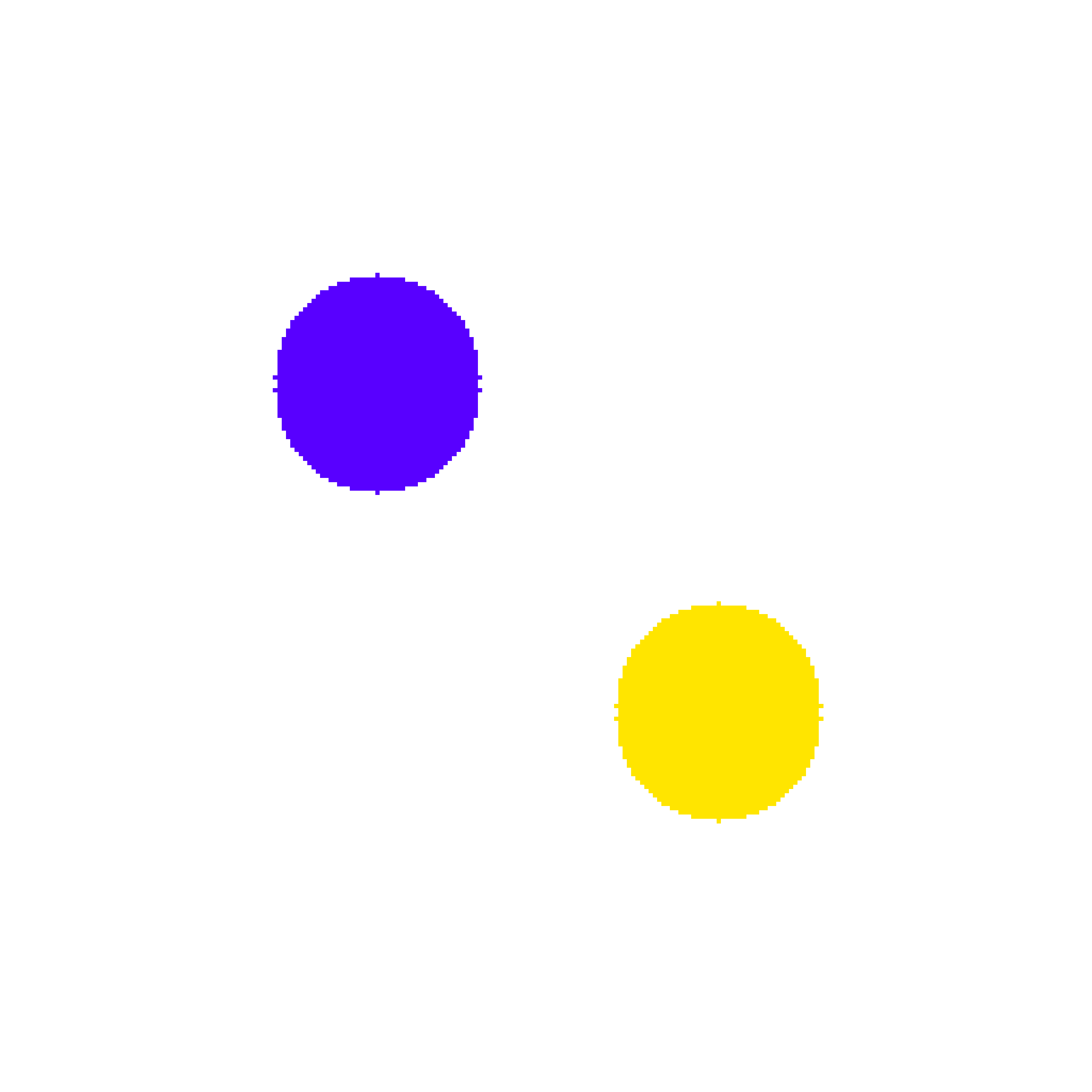}
    \includegraphics[width=0.23\textwidth]{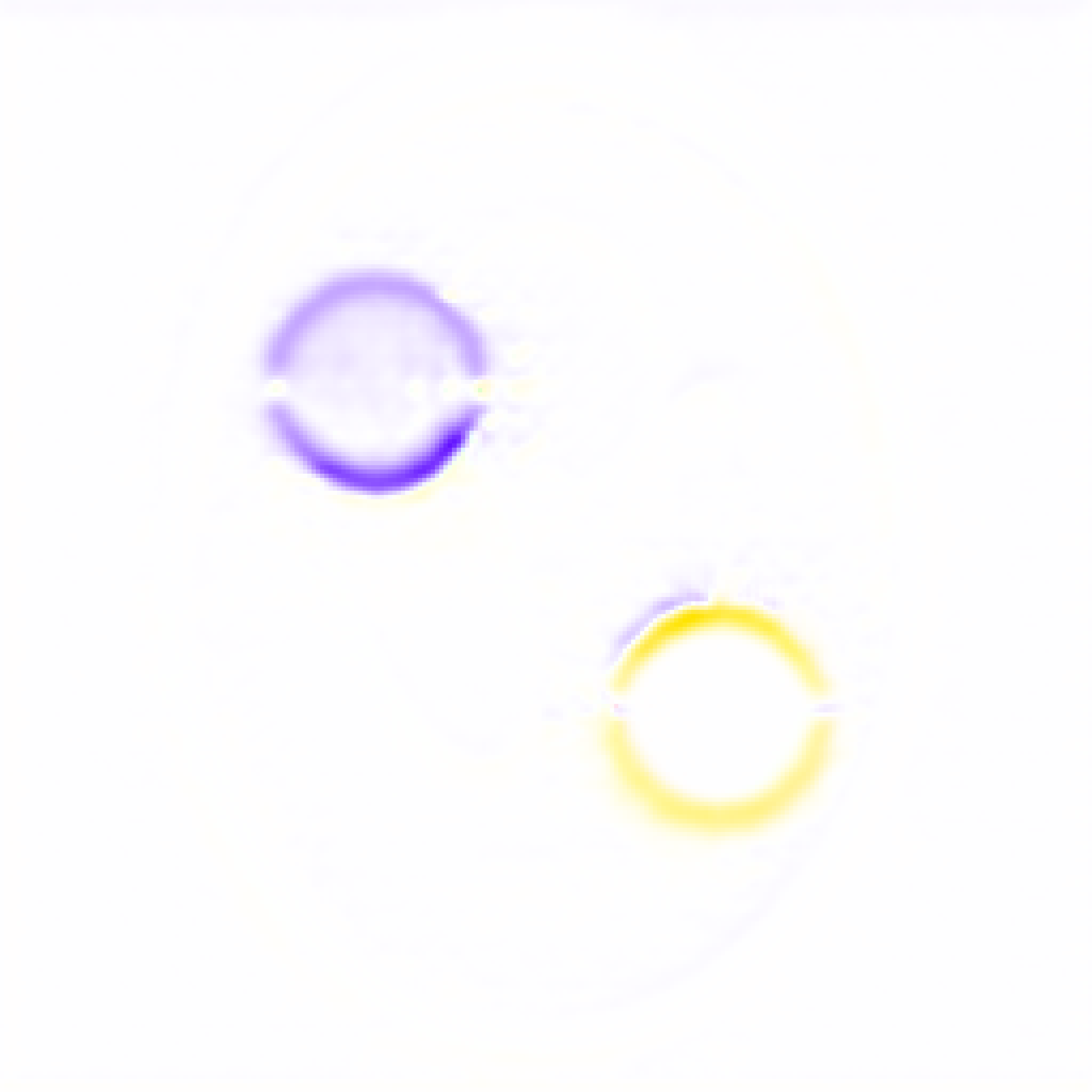}
    \includegraphics[width=0.23\textwidth]{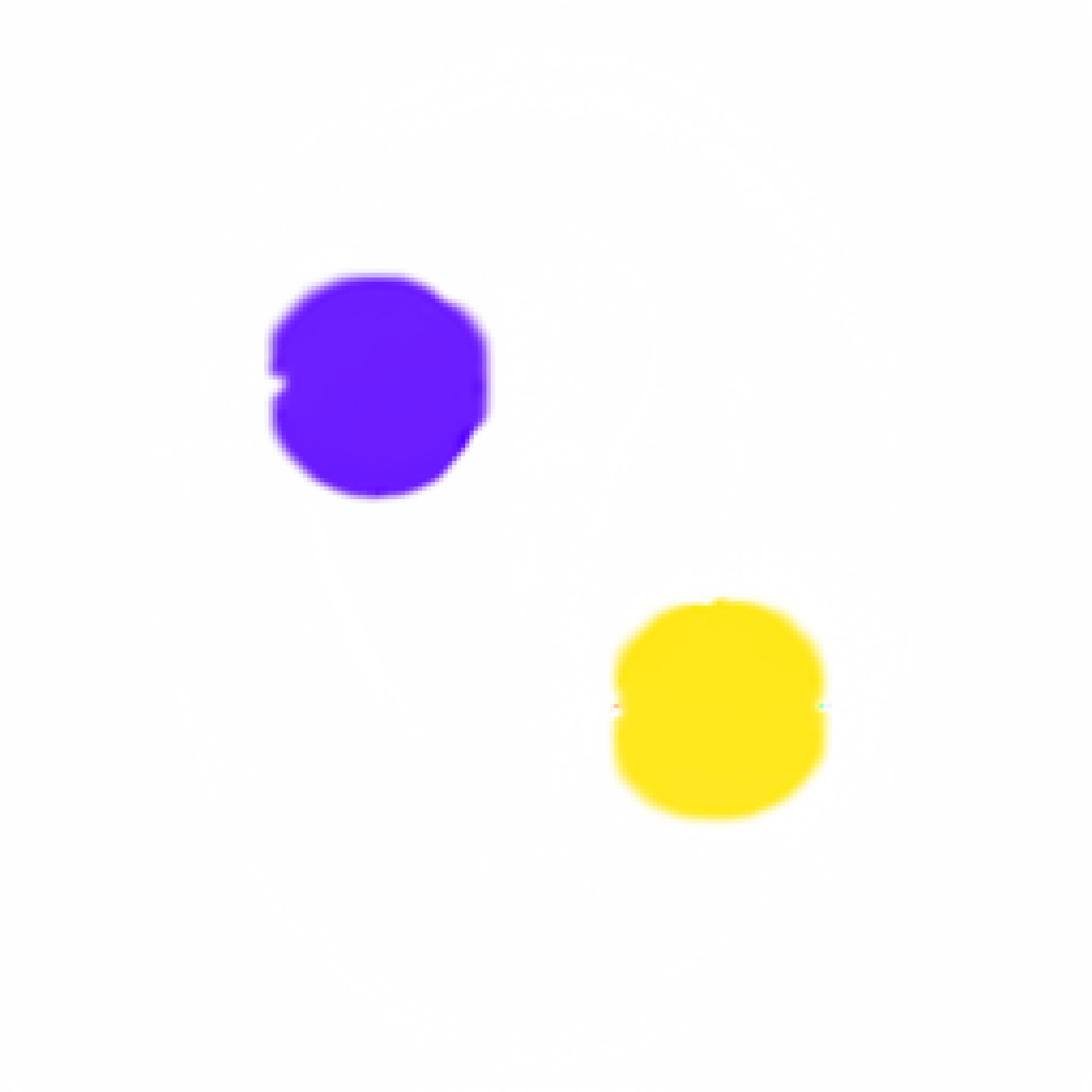}
    \caption{\small Synthetic Shepp-Logan phantom experiment. Top row (left to right): original phantom, synthetic frame 1 with embedded circles, warped frame 2, and color wheel. Bottom row: ground truth flow, predicted flows for $\lambda_{\mathrm{TV}} = 0$ and $10^{-5}$, respectively, all trained for 10{,}000 epochs. 
    }
\label{fig:shepp-logan}
\end{figure}

Each circular region is assigned opposing vertical motion to simulate anatomical displacement patterns: the upper circle is translated upward by 3 pixels while the lower circle is shifted downward by the same amount. 
We warp the original image according to this synthetic flow field to generate the target frame.
The obtained image frames and the respective ground-truth optical flow fields are depicted in \cref{fig:shepp-logan}.

All flow vectors are normalized by the maximum displacement magnitude to ensure optimal color saturation, allowing regions with maximum motion to appear as fully saturated colors. This synthetic setup provides an idealized benchmark for evaluating unsupervised optical flow models, combining well-defined anatomical structure with ground-truth motion fields. 

We train the model for a fixed amount of 10{,}000 epochs, which allows the model to achieve highly accurate reconstruction of the imposed deformation, using two separate runs with two different TV weights, $\lambda_\TV=0$ and $\lambda_\TV = 10^{-5}$. \Cref{fig:shepp-logan} illustrates the predicted flow for both choices of $\lambda_\TV$ and demonstrates that the learned network
captures the opposite vertical displacement of the two circles and shows increased smoothness with regularization.

\begin{figure}[H]
    \centering
    \begin{adjustbox}{center}
    \includegraphics[width=0.33\textwidth]{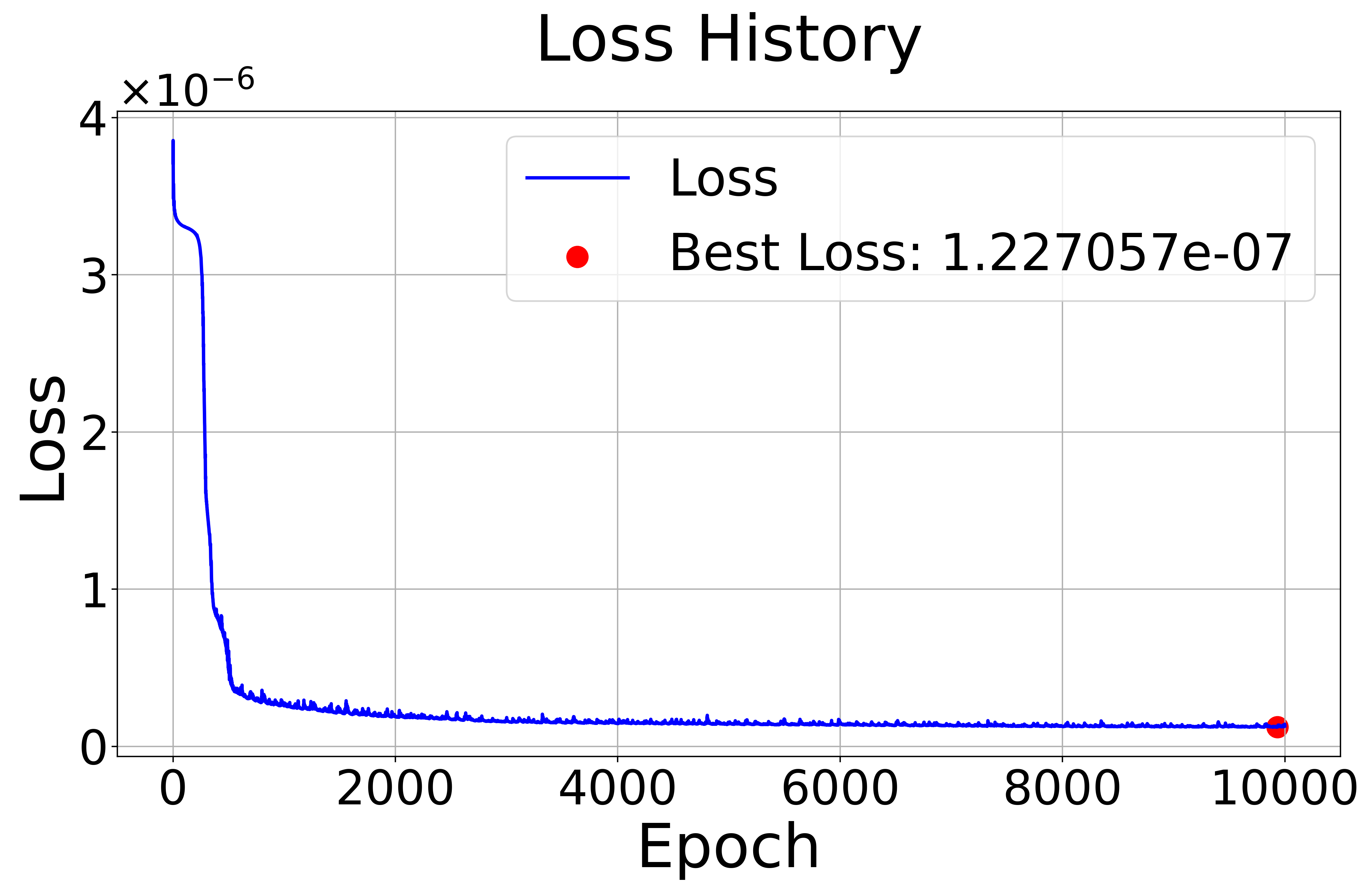}
    \includegraphics[width=0.33\textwidth]{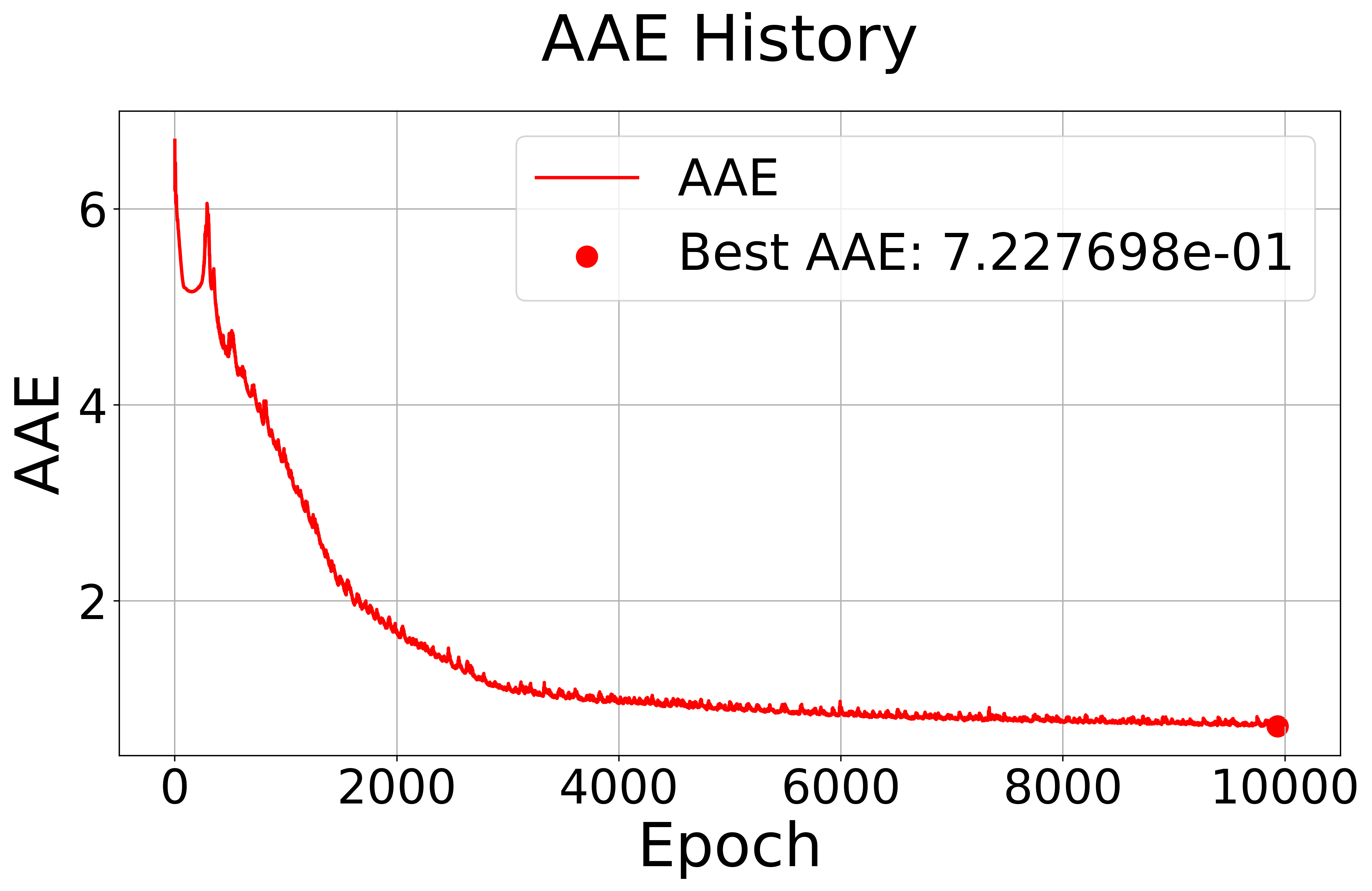}
    \includegraphics[width=0.33\textwidth]{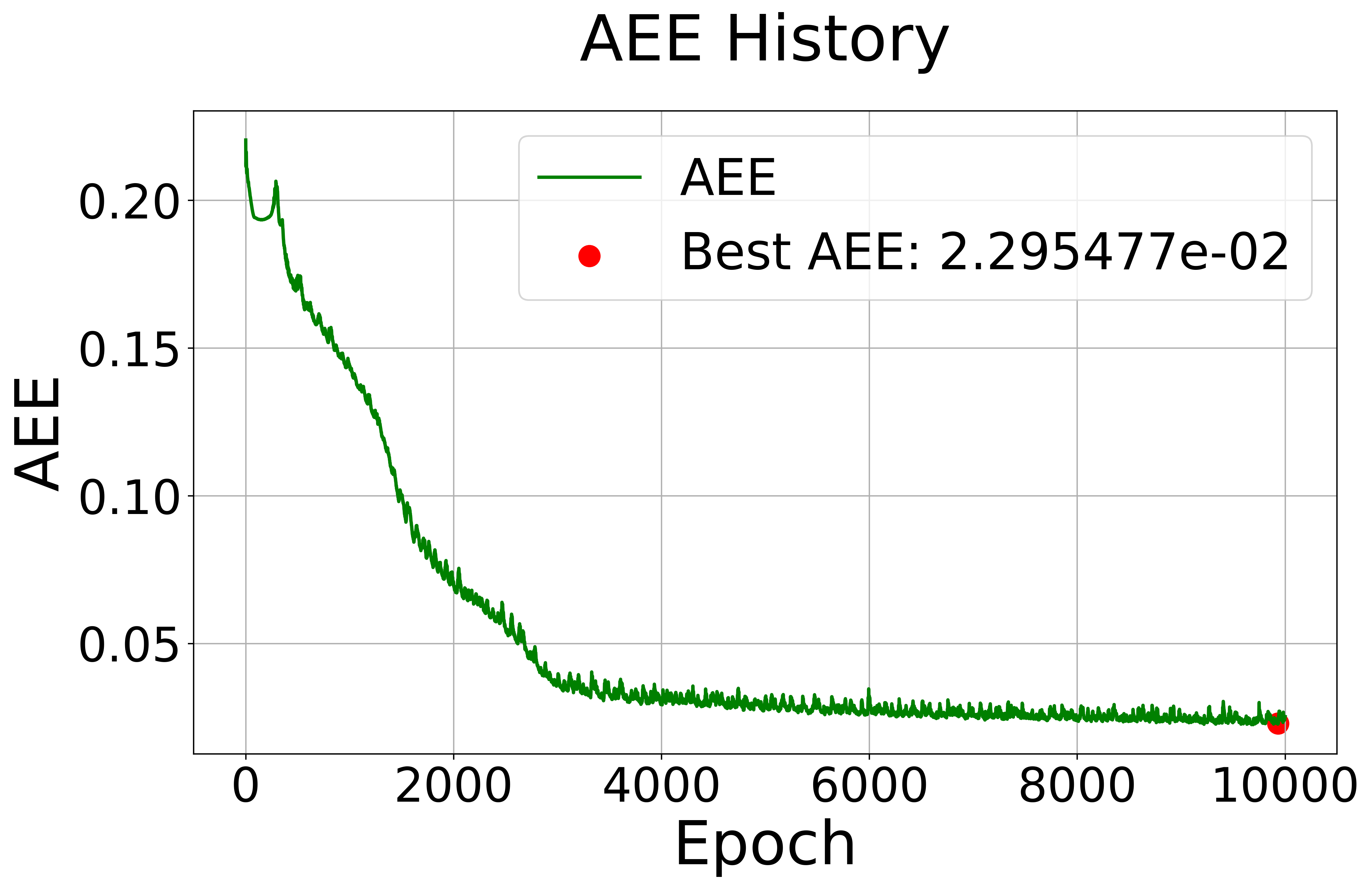}
    \end{adjustbox}
\caption{\small Training curves for the Shepp-Logan phantom experiment using total variation regularization with $\lambda_{\mathrm{TV}} = 10^{-5}$. The model is trained for 10{,}000 epochs. \text{Left:} Loss history showing stable convergence with a final best loss of $1.23 \times 10^{-7}$. \text{Middle:} AEE curve indicating accurate magnitude estimation of flow vectors, with a best AEE of $2.30 \times 10^{-2}$ and SDEE of $1.88 \times 10^{-1}$. \text{Right:} AAE decreasing to a final value of $7.23\times 10^{-1}$, with SDAE of $5.87$.}
\label{fig:shepp-logan-metrics}
\end{figure}

In \cref{fig:shepp-logan-metrics}, the training dynamics for $\lambda_\TV = 10^{-5}$ are summarized. Specifically, we observe that the loss curve (left) shows a rapid decrease within the first 1,000 epochs and continues to decline more gradually thereafter, indicating successful reconstruction of the synthetic motion patterns. AEE (middle) demonstrates an accurate magnitude estimation of flow vectors, while AAE (right) shows precise directional learning with consistent improvement throughout training. These metrics confirm that the network successfully learns both accurate magnitude and directional representations of the opposing circular motions. Note that the final best loss is $1.23 \times 10^{-7}$, indicating that the constant brightness assumption \eqref{Eq:ConstantBrightness} is closely satisfied by the estimated optical flow.

Although the stopping criterion is a fixed number of epochs, the downward trends in all error curves of \cref{fig:shepp-logan-metrics} suggest that additional training could yield further improvements. The use of total variation regularization contributes to the smoothness and sharpness of the estimated flow field, particularly in preserving boundaries of the moving structures. Together, these results validate the model’s ability to resolve localized motion in an interpretable, noise-free environment.

\subsection{Results on Middlebury Benchmark}

We evaluate FractalPINN-Flow on the Middlebury optical flow benchmark~\cite{baker2011database}  to assess both quantitative accuracy and visual quality across a range of TV weights $\lambda_\TV \in \{0, 10^{-3}, 10^{-2}, 10^{-1}\}$. All models are trained for 20{,}000 epochs with fixed fractal depth $d = 4$. Table~\ref{tab:tv-results} reports the best training loss, epoch of best performance, and evaluation metrics—AEE, SDEE, AAE, and SDAE—for each benchmark scene. 
Across most benchmarks, intermediate regularization values ($\lambda_{\mathrm{TV}} = 10^{-2}$ or $10^{-3}$) tend to yield the lowest AEE and AAE, striking a favorable balance between motion detail preservation and flow smoothness.
For instance, in the \textit{Dimetrodon} and \textit{RubberWhale} scenes, $\lambda_{\mathrm{TV}} = 10^{-2}$ achieves the lowest AEE (0.33 and 0.17, respectively). Similarly, \textit{Venus} shows optimal performance at $\lambda_{\mathrm{TV}} = 10^{-2}$ (AEE = 0.31, AAE = 0.08), while \textit{Hydrangea} reaches its best results at the same setting (AEE = 0.43, AAE = 0.12). However, exceptions are present: in the \textit{Grove3} scene, the best AEE (1.16) and AAE (0.17) are obtained at $\lambda_{\mathrm{TV}} = 10^{-1}$, outperforming lower regularization levels. In contrast, strongly regularized configurations degrade performance in noisy or high-disparity settings such as \textit{Urban2}, where $\lambda_{\mathrm{TV}} = 10^{-1}$ yields an AEE of 7.64 versus 2.61 at $\lambda_{\mathrm{TV}} = 10^{-2}$. These results emphasize the critical role of appropriately tuned regularization in guiding unsupervised optical flow, particularly in complex scenes with variable texture and motion.

\begin{table}[h]
\centering

\begin{tabular}{cccccccc}
\toprule
Benchmark & $\lambda_{\mathrm{TV}}$ & Best Loss & Best Epoch & AEE & SDEE & AAE & SDAE \\
\midrule
Dimetrodon & 0 & 0.000243 & 19867 & 0.77 & 0.63 & 0.31 & 0.31 \\
 & ${10^{-1}}$ & 0.001957 & 19625 & 0.42 & 0.46 & 0.16 & 0.23 \\
 & ${10^{-2}}$ & 0.001187 & 19928 & 0.33 & 0.47 & 0.13 & 0.23 \\
 & ${10^{-3}}$ & 0.000726 & 19794 & 0.41 & 0.47 & 0.17 & 0.24 \\
Grove2 & 0 & 0.000822 & 19648 & 0.52 & 0.55 & 0.12 & 0.16 \\
 & ${10^{-1}}$ & 0.006699 & 19284 & 1.18 & 1.54 & 0.52 & 0.73 \\
 & ${10^{-2}}$ & 0.002833 & 19972 & 0.2 & 0.42 & 0.05 & 0.12 \\
 & ${10^{-3}}$ & 0.001276 & 19610 & 0.34 & 0.43 & 0.08 & 0.12 \\
Grove3 & 0 & 0.001730 & 19779 & 1.94 & 2.31 & 0.45 & 0.58 \\
 & ${10^{-1}}$ & 0.011981 & 19866 & 1.16 & 1.67 & 0.17 & 0.31 \\
 & ${10^{-2}}$ & 0.004666 & 19292 & 1.18 & 1.87 & 0.2 & 0.33 \\
 & ${10^{-3}}$ & 0.002215 & 19914 & 1.34 & 1.97 & 0.26 & 0.39 \\
Hydrangea & 0 & 0.000414 & 19749 & 0.74 & 1.25 & 0.18 & 0.34 \\
 & ${10^{-1}}$ & 0.005052 & 19867 & 0.48 & 1.04 & 0.13 & 0.32 \\
 & ${10^{-2}}$ & 0.001893 & 19964 & 0.43 & 1.12 & 0.12 & 0.32 \\
 & ${10^{-3}}$ & 0.000810 & 19773 & 0.56 & 1.21 & 0.15 & 0.33 \\
RubberWhale & 0 & 0.000190 & 19822 & 0.43 & 0.53 & 0.22 & 0.27 \\
 & ${10^{-1}}$ & 0.002276 & 19969 & 0.34 & 0.6 & 0.19 & 0.35 \\
 & ${10^{-2}}$ & 0.001068 & 19914 & 0.17 & 0.38 & 0.1 & 0.24 \\
 & ${10^{-3}}$ & 0.000488 & 19722 & 0.25 & 0.42 & 0.13 & 0.24 \\
Urban2 & 0 & 0.001047 & 19639 & 3.5 & 5.64 & 0.28 & 0.42 \\
 & ${10^{-1}}$ & 0.010732 & 19757 & 7.64 & 7.77 & 0.75 & 0.43 \\
 & ${10^{-2}}$ & 0.003046 & 19877 & 2.61 & 5.06 & 0.11 & 0.23 \\
 & ${10^{-3}}$ & 0.001351 & 19579 & 2.74 & 5.26 & 0.14 & 0.29 \\
Urban3 & 0 & 0.000675 & 19779 & 3.4 & 4.47 & 0.38 & 0.7 \\
 & ${10^{-1}}$ & 0.007248 & 19541 & 4.68 & 4.28 & 0.47 & 0.72 \\
 & ${10^{-2}}$ & 0.003074 & 19637 & 3.26 & 3.95 & 0.33 & 0.63 \\
 & ${10^{-3}}$ & 0.001127 & 19331 & 2.6 & 4.17 & 0.3 & 0.7 \\
Venus & 0 & 0.000583 & 19828 & 0.73 & 0.89 & 0.17 & 0.33 \\
 & ${10^{-1}}$ & 0.009652 & 19799 & 1.7 & 1.96 & 0.62 & 0.7 \\
 & ${10^{-2}}$ & 0.001779 & 19646 & 0.31 & 0.66 & 0.08 & 0.29 \\
 & ${10^{-3}}$ & 0.000997 & 19828 & 0.46 & 0.92 & 0.11 & 0.32 \\
\bottomrule
\end{tabular}
\caption{\small Benchmark results for various $\lambda_{\mathrm{TV}}$ configurations, based on training for 20{,}000 epochs.}
\label{tab:tv-results}
\end{table} 

\begin{figure}[H]
    \centering
    \begin{adjustbox}{center}
        \includegraphics[width=1.01\textwidth]{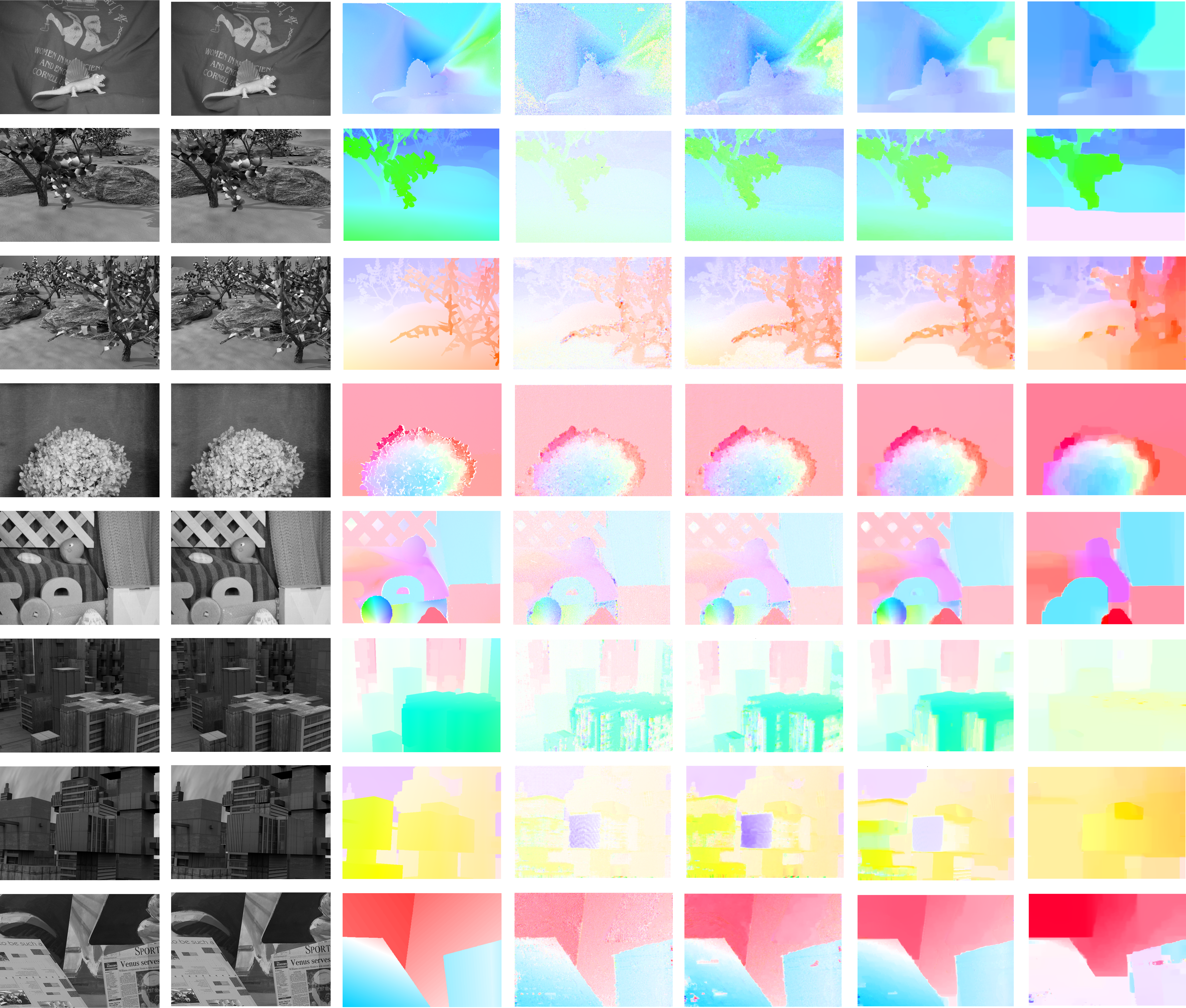}
    \end{adjustbox}
\caption{\small Middlebury Optical Flow Benchmark visualizations corresponding to the results in Table~\ref{tab:tv-results}. Columns from left to right: $I_1$, $I_2$, ground truth optical flow, and predicted flows for $\lambda_{\mathrm{TV}} = 0$, $10^{-3}$, $10^{-2}$, $10^{-1}$. Predicted flow fields are taken from the best-loss epoch for each configuration. Benchmarks from top to bottom: \textit{Dimetrodon}, \textit{Grove2}, \textit{Grove3}, \textit{Hydrangea}, \textit{RubberWhale}, \textit{Urban2}, \textit{Urban3}, \textit{Venus}.}
    \label{fig:flow-visualization}
\end{figure}

Figure~\ref{fig:flow-visualization} visualizes the predicted flow fields for each configuration, revealing the qualitative effects of $\lambda_{\mathrm{TV}}$ on spatial smoothness and edge preservation. High regularization improves visual coherence but risks oversmoothing fine structures, while low or zero regularization retains motion discontinuities but introduces noise and instability. These findings demonstrate the capacity of our fractal-based model to generalize across a wide spectrum of motion patterns and visual complexities, while also emphasizing the practical importance of hyperparameter tuning in unsupervised flow models.

The results in Table~\ref{tab:tv-results} highlight the sensitivity of FractalPINN-Flow to the choice of total variation regularization weight $\lambda_{\mathrm{TV}}$. Across most benchmarks, introducing moderate regularization ($\lambda_{\mathrm{TV}} = 10^{-2}$) consistently yields the lowest AEE and AAE, suggesting that total variation plays a key role in suppressing noise while preserving meaningful motion boundaries.

\bibliographystyle{plain}
\section*{Code availability}
The full implementation, training scripts, and experiment configurations for \textit{FractalPINN-Flow} are available at
\href{https://github.com/sarabehnamian/FractalPINN-Flow}{github.com/sarabehnamian/FractalPINN-Flow}.

\bibliography{references}

\end{document}